\documentclass[letterpaper, 10 pt, conference]{ieeeconf}
\IEEEoverridecommandlockouts                  
\usepackage{amsmath,amsfonts,amssymb}
\usepackage{graphicx}
\usepackage{booktabs}
\usepackage{arydshln}
\usepackage{CJKutf8}

\usepackage[colorlinks=true, allcolors=blue]{hyperref}
\usepackage{multirow}

\usepackage{color}
\usepackage{booktabs}

\usepackage{xcolor}

\title{\LARGE \bf
{SOS-1K}: A Fine-grained Suicide Risk Classification Dataset for Chinese Social Media Analysis}

\author{Hongzhi Qi$^{1}$, Hanfei Liu$^{1}$, Jianqiang Li$^{1}$, Qing Zhao$^{1}$, Wei Zhai$^{1}$, Dan Luo$^{2}$, \\Tian Yu He$^{2}$,  Shuo Liu$^{2}$,  Bing Xiang Yang$^{2}$ and Guanghui Fu\textsuperscript{*}$^{3}$
\thanks{*This work was supported by grants from the National Natural Science Foundation of China (grant numbers:72174152, 72304212 and 82071546), Fundamental Research Funds for the Central Universities (grant numbers: 2042022kf1218; 2042022kf1037), the Young Top-notch Talent Cultivation Program of Hubei Province. Corresponding author: Guanghui Fu (\url{guanghui.fu@inria.fr})}
\thanks{$^{1}$School of Software Engineering, Beijing University of Technology, Beijing, China}%
\thanks{$^{2}$School of Nursing, Wuhan University, Wuhan, China}%
\thanks{$^{3}$Sorbonne Université, Institut du Cerveau - Paris Brain Institute - ICM, CNRS, Inria, Inserm, AP-HP, Hôpital de la Pitié Salpêtrière, Paris, France}%
}

\begin{document}

\maketitle
\pagestyle{plain}

\begin{abstract}

In the social media, users frequently express personal emotions, a subset of which may indicate potential suicidal tendencies. The implicit and varied forms of expression in internet language complicate accurate and rapid identification of suicidal intent on social media, thus creating challenges for timely intervention efforts.
The development of deep learning models for suicide risk detection is a promising solution, but there is a notable lack of relevant datasets, especially in the Chinese context. To address this gap, this study presents a Chinese social media dataset designed for fine-grained suicide risk classification, focusing on indicators such as expressions of suicide intent, methods of suicide, and urgency of timing. Seven Pre-trained models were evaluated in two tasks: high and low suicide risk, and fine-grained suicide risk classification on a level of 0 to 10. In our experiments, deep learning models show good performance in distinguishing between high and low suicide risk, with the best model achieving an F1 score of 88.39\%. However, the results for fine-grained suicide risk classification were still unsatisfactory, with an weighted F1 score of 50.89\%. To address the issues of data imbalance and limited dataset size, we investigated both traditional and advanced, large language model based data augmentation techniques, demonstrating that data augmentation can enhance model performance by up to 4.65\% points in F1-score. Notably, the Chinese MentalBERT model, which was pre-trained on psychological domain data, shows superior performance in both tasks. This study provides valuable insights for automatic identification of suicidal individuals, facilitating timely psychological intervention on social media platforms. The source code and data are publicly available at: \url{https://github.com/HongzhiQ/FineGrainedSuicideDetection}.
\end{abstract}

\section{Introduction}
\label{sec:intro} 

Mental health disorders have emerged as a significant public health challenge worldwide. A report from the World Health Organization (WHO) indicates that one in eight individuals globally suffers from mental disorders~\cite{world2022world}. Severe mental illnesses often precipitate suicidal behaviors. With the advent of the internet, social media platforms have risen rapidly, becoming a refuge for many patients who share their distress and seek emotional support~\cite{keles2020systematic}. On these platforms, a significant number of users express their existential pain under specific topics, with some even disclosing suicidal tendencies~\cite{chen2020treehole}. Monitoring mental health through social media content enables early diagnosis before the deterioration of a user's psychological state, thus potentially averting the occurrence of suicide~\cite{yang2021suicide}. 
In China, the Tree Hole Rescue Group harnessed AI to detect potential online suicide cases, successfully intervening in 3,629 instances with the assistance of 600 volunteers~\cite{yang2021suicide, yimeng2020aigives}. 
Given the high volume of posts on social media, manual review methods often fail to monitor and assess the vast amount of data in time, posing a significant challenge to the timely intervention of emergencies such as suicide, and hence there is an urgent need for an accurate and efficient automated identification method.

Deep learning methods presents a promising avenue for addressing this challenge~\cite{coppersmith2018natural}.
While existing studies achieve good classification results, the granularity of the classification is relatively coarse~\cite{fu2021distant,renjith2022ensemble,gorai2024bert}. To address this issue, Wang et al.~\cite{wang2021medical} proposed a deep evaluation model based on knowledge-perception that integrates the prediction results based on a suicide dictionary with those from the BERT model, ultimately achieving a F1 score of 54\% on a time-based 10-category dataset. However, this method requires the construction of a large, specialized suicide dictionary, a process that is time-consuming and labor-intensive, and prone to the out-of-vocabulary problem. Furthermore, classification tasks also face the issue of class imbalance, for which data augmentation is an effective solution~\cite{shorten2021text}. Yet, the specific methods that effectively enhance model performance require further investigation~\cite{10.1162/tacl_a_00542}.

The advent of large language models (LLMs) has further accelerated technological transformation in the field of mental health~\cite{zhao2023survey, kaddour2023challenges,he2023towards}. 
Numerous studies are dedicated to leveraging LLMs for interventions in psychological issues~\cite{liu2023chatcounselor, fu2023enhancing, zhou2023identifying, ayers2023comparing}. 
Qi et al.~\cite{qi2023supervised} conducted experiments comparing the performance of LLMs and supervised learning models on mental health tasks in Chinese social media, with a focus on the identification of cognitive distortions and the prediction of high and low suicide risk. However, the experimental results indicated that LLMs still lag behind supervised learning methods in terms of effectiveness in mental health tasks. This suggests that supervised learning approaches remain the preferred solution for addressing mental health challenges.

To address this gap, we introduce a new dataset name as {SOS-1K}, which contains suicide-related data collected from Chinese social media platforms. This dataset has been classified into 11 levels based on factors such as the expression of existential pain, the intensity of suicidal intention, methods of suicide, and the definiteness and urgency of suicide plans, encompassing a total of 1249 entries. 
Additionally, we simplified the fine-grained classification task into a binary high-low risk classification to meet different practical needs.

In this study, we evaluated the performance of seven pre-trained models on both fine-grained and high-low suicide classification tasks using our proposed dataset. We also explored various data augmentation techniques to address category imbalance in the fine-grained task, including traditional and LLM based methods. 
Utilizing data augmentation for the fine-grained task, the Chinese MentalBERT~\cite{zhai2024chinese} model achieving 55.54\% in the fine-grained task and 88.39\% in the high-low tasks, respectively.
The proposed dataset, along with publicly available models, has potential for detecting suicide-related content on social media platforms.

\section{Task definition} \label{sec:task_definition}

\subsection{Fine-grained suicide risk classification}\label{sec:annotation:fine_grained}

Suicide does not occur abrupty and without warning. In fact, most cases follow certain patterns~\cite{xiao2024decoding}. Therefore, analyzing and assessing relevant factors can aid in predicting users' suicidal behavior and enable timely intervention~\cite{cowen2012shorter}. Huang et al.~\cite{Huang2019WebBasedIA} identified five key factors: existential distress, suicidal intent, suicide plan, method of suicide, and date of suicide. Based on factors such as the expression of existential distress, the intensity of suicidal intent, the method of suicide, the certainty of the suicide plan, and the urgency of time, a 0-10 level suicide risk classification label was determined. Details of the fine-grained suicide risk classification are presented in Table~\ref{tab:suicide_huang_rule}. 

\begin{table*}
\caption{The 0-10 suicide risk classification rule defined by Huang et al.~\cite{Huang2019WebBasedIA}}
\centering
\resizebox{0.8\linewidth}{!}{
\begin{tabular}{ll} 
\bottomrule
\textbf{Level} & \multicolumn{1}{c}{\textbf{Description}}                                                          \\ 
\bottomrule
level 0~ ~     & No expression of existential pain.                                                                \\
level 1~ ~     & Existential pain is expressed, with no expression of suicidal desire.                             \\
level 2~       & Existential pain has been clearly expressed, with no expression of suicidal desire.               \\
level 3~~      & Strong existential pain, with no expression of suicidal desire.                                   \\
level 4~ ~     & Suicidal desire has been expressed, but the method and plan for suicide are unclear.              \\
level 5~ ~     & Strong desire for suicide, with an unclear method of suicide.                                     \\
level 6~ ~     & Planning for suicide has begun, but the method of suicide is unknown.                             \\
level 7~ ~     & The method of suicide has been determined, but the date of suicide has not.                       \\
level 8~ ~     & Suicide has been planned, with the date of suicide roughly determined.                            \\
level 9~ ~     & The method of suicide has been determined, and the person may commit suicide in the near future.  \\
level 10~ ~    & Suicide may be ongoing.                                                                           \\
\bottomrule
\end{tabular}
}
\label{tab:suicide_huang_rule}
\end{table*}

\subsection{High-low suicide risk classification}\label{sec:annotation:high_low}

As mentioned above, a fine-grained classification of suicide risk is crucial. However, for practical applications, simplifying the classification into high and low risk can be effective. For individuals identified as experiencing a moderate emotional crisis (categorized as low risk), psychological interventions can be effectively used to manage and alleviate their emotional distress. In contrast, individuals classified as high risk require immediate and specialized crisis intervention strategies due to the urgency of their situation.  Therefore, we classify levels 6 and above as high risk, and levels 1 to 5 as low risk.

\section{Methods} \label{sec:methods}

\begin{itemize}
    \item \textbf{BERT~\cite{kenton2019bert}:} Utilizes the Transformer's~\cite{vaswani2017attention} encoder and bidirectional processing to acquire contextual word representations. Its core pre-training involves the Masked Language Model (MLM), where it predicts randomly hidden words in a sentence, and Next Sentence Prediction (NSP), assessing if two sentences are sequential. These tasks collectively refine BERT's grasp of complex linguistic nuances.
    \item \textbf{RoBERTa~\cite{liu2019roberta}:} Improves upon BERT by removing the NSP task, using a larger corpus, and augmenting batch sizes and training steps. It also updates BERT's dynamic masking for data variation in each training iteration. These enhancements boost RoBERTa's performance in numerous NLP tasks.
     \item \textbf{ELECTRA~\cite{clark2020electra}:} Introduces replaced token detection mechanism, diverging from the conventional MLM approach. Instead of predicting masked words, ELECTRA uses a generator to predict substitutions for ``[MASK]'' tokens in sentences and then employs a discriminator to detect these replacements. This method enhances computational efficiency by utilizing all input words for training, not just the masked ones.
    \item \textbf{MacBERT~\cite{cui2021pre}:} Enhances the BERT architecture with a refined MLM task and curricular learning strategies, incorporating whole word masking for better suitability with Chinese texts. This advanced MLM task involves predicting masked words and identifying inaccuracies, catering to complex language scenarios. These improvements enable MacBERT to outperform BERT in various natural language processing tasks.

    \item \textbf{NeZha~\cite{wei2019nezha}:} Enhances Transformer~\cite{vaswani2017attention} models with functional relative position encoding for long sequences, improving sensitivity to context through long-distance dependency capture. It includes whole-word masking and mixed-precision training, achieving superior performance in Chinese language tasks.
     \item \textbf{ERNIE 3.0~\cite{sun2021ernie}:} Enhances the ERNIE series by combining language understanding and generation tasks with a heterogeneous knowledge graph for deeper semantic insights. It uses graph attention for entity and relationship integration and employs multi-task learning, including traditional MLM and knowledge-driven tasks, improving domain-specific performance.
    \item \textbf{Chinese MentalBERT~\cite{zhai2024chinese}:} A domain-adaptive model for the Chinese mental health field, utilizing 3.36 million pieces of data, this model employs a lexicon-guided masking mechanism during the pre-training phase, prioritizing the masking of keywords related to mental health. It excels in mental health tasks, effectively capturing psychological nuances.

\end{itemize}

\section{Data augmentation}\label{sec:methods:data_augmentation}

In our study on fine-grained suicide risk classification, we face the challenge of class imbalance. We employed data augmentation methods to improve the model's ability to generalize. 
We use two traditional data augmentation methods that proved to be effective~\cite{chen2023empirical}: synonym replacement~\cite{kolomiyets2011model} for token-level augmentation and round-trip translation~\cite{aiken2010efficacy} for sentence-level augmentation.
Additionally, we employed a novel data augmentation method that utilizes LLM with few-shot training examples as prompts. 

\subsection{Augmentation methods}
\begin{itemize}
    \item Synonym replacement~\cite{kolomiyets2011model}: It is a token-level augmentation method that boosts the diversity of a corpus by replacing words or phrases with their synonyms. In our experiment, the TF-IDF algorithm~\cite{salton1975vector} was utilized to identify the keywords in each text, preserving it while replacing the remaining vocabulary with synonyms selected from the \texttt{synonyms} Python package~\cite{Synonyms:hain2017} based on word embedding similarity. For each keyword, the top four related synonyms were chosen to generate four innovative variant texts for every original text.
    \item Round-trip translation~\cite{aiken2010efficacy}: It is a sentence-level augmentation method used to create text variations by leveraging translation tools. This method involves translating a set of original texts from one language to another, and then translating these texts back to the original language. In our implementation, we chose English, Spanish, Russian, Arabic, and Japanese as the languages for translation. We used the Baidu Translation API to facilitate this process~\cite{BaiduTranslateAPI} generating five new textual variations for each original text. 
    \item Data Generation by LLM: We utilized the GPT-4 model~\cite{achiam2023gpt} with few-shot prompting: randomly chose six expert-annotated sentences from each category to serve as few shot training data. The temperature was set to 0.7. We generated 350 candidate data entries for each level and randomly selected them according to a specific data volume strategy.
\end{itemize}

\subsection{Augmentation volumes}

To address class imbalance, we used two strategies to ensure that each of the categories achieves a predefined augmentation volume. Let's define $L$ as the total number of categories. For each category $l, N_l$ denotes its original data size. Additionally, $N_l^a$ represents the target data volume for category $l$ after applying a specific strategy.
\begin{itemize}
    \item \textbf{Maximal category augmentation}: To address the problem of inter-class imbalance, this strategy adjusts the target augmentation volume to match the size of the largest category. The target volume $N_l^a$ is:
    $N_l^a = \max_{l=1}^{L} N_l$
    \item \textbf{Double maximal category augmentation}: This strategy doubles the augmentation volume to test whether increasing the data volume can enhance model performance by providing a richer and more varied dataset for training. Specifically, the target augmentation volume for each class is set to $2 \times N_l^{a}$.
\end{itemize}

\section{Experiments} \label{sec:experiments}

\subsection{Datasets}

We obtained data by scraping comments from the ``Zoufan''~\cite{zoufan} blog on the Weibo social platform. 
The initial annotation of the data was conducted using LLM, followed by a secondary review and annotation by domain experts (B.Y). This process led to the conduction of the SOcial Media Suicide dataset ({SOS-1k}), which contains 1,249 entries. Details of the annotation rule are described in Section~\ref{sec:task_definition}.
The data obtained is publicly available on Weibo, and we remove all information such as user IDs for anonymization to ensure privacy protection.
Data statistics for the fine-grained and high-low suicide risk classification tasks in the {SOS-1K} dataset are shown in Table~\ref{tab:dataset}. 
For both tasks, we use a 4:1 ratio to split the data into training and testing sets. Additionally, we conduct 5-fold cross-validation to ensure the robustness of our models.

\begin{table}
\centering
\caption{Distribution of the {SOS-1k} dataset. $N_l$ represents the number of samples at level $l$. $N_{\text{train}}$ and $N_{\text{test}}$ represent the number of data samples in train and test set, respectively. $\overline{W}$ denotes the average word count per sample. }
\label{tab:dataset}
\begin{tabular}{|c|c|c|} 
\hline
Level $l$& Risk ($N_{\text{risk}}$)                            & $N_{l}$                                                                                 \\ 
\hline
0     & \multirow{6}{*}{Low (648)}  & 64                                                                                    \\ 
\cline{1-1}\cline{3-3}
1     &                                  & 63                                                                                    \\ 
\cline{1-1}\cline{3-3}
2     &                                  & 67                                                                                    \\ 
\cline{1-1}\cline{3-3}
3     &                                  & 148                                                                                   \\ 
\cline{1-1}\cline{3-3}
4     &                                  & 248                                                                                   \\ 
\cline{1-1}\cline{3-3}
5     &                                  & 93                                                                                    \\ 
\hline
6     & \multirow{5}{*}{High (601)} & 213                                                                                   \\ 
\cline{1-1}\cline{3-3}
7     &                                  & 199                                                                                   \\ 
\cline{1-1}\cline{3-3}
8     &                                  & 41                                                                                    \\ 
\cline{1-1}\cline{3-3}
9     &                                  & 71                                                                                    \\ 
\cline{1-1}\cline{3-3}
10    &                                  & 42                                                                                    \\ 
\hline
\multicolumn{3}{|c|}{$N_{\text{train}}$ = 999; $N_{\text{test}}$ = 250; $N_{\text{total}}=1249$; $\overline{W}$=47.79}    \\
\hline
\end{tabular}
\end{table}

\subsection{Implementation details}
\label{sec:Experiment design}

This study evaluates the performance of seven pre-trained language models on two suicide risk classification tasks. To address dataset imbalance in the fine-grained suicide classification task, weighted average precision, recall, and F1-score were used as evaluation metrics. For the high-low classification task, binary averages were employed.

Experiments were conducted utilizing the PyTorch~\cite{paszke2019pytorch} framework.
The Adam optimizer~\cite{kingma2015adam} was utilized for training. The learning rate and batch size vary according to the model and the amount of training data. 
All text data were uniformly padded or truncated to 150 tokens for preprocessing.
A fully connected layer with a sigmoid activation function was introduced after the pre-trained model to serve as a classifier. To prevent overfitting, an early stopping mechanism was employed; model training will stop if there is no decrease in loss on the validation set for 30 consecutive epochs. All the model settings, data and codes are publicly available via: \url{https://github.com/HongzhiQ/FineGrainedSuicideDetection}.

\section{Results} 
\label{sec:results}

\subsection{The performance of suicide risk classification tasks}
The performance of the pre-trained models on the fine-grained and high-low suicide risk classification tasks is shown in Table~\ref{tab:result}.

\subsubsection{Fine-grained multi-class suicide risk classification}
All models failed to achieve high F1 scores, with the top-performing model, Chinese MentalBERT~\cite{zhai2024chinese}, reaching only 50.89\%. This may be due to the task's numerous categories and the inherent difficulty in distinguishing between their respective semantic nuances. 
The relative outperformance of the Chinese MentalBERT~\cite{zhai2024chinese} model compared to other pre-trained models may be due to its training for domain adaptation on a large corpus of psychology-related social media content. This specialized training enhanced its ability to understand and process the specific language and contexts of this field. Note that, the MacBERT~\cite{cui2021pre} closely follows, trailing by just 0.3\% points in F1-score for this task.

\subsubsection{High-low suicide risk binary classification}
The experimental results demonstrate that all models performed well in this task, with F1-scores exceeding 80\%. Notably, the Chinese MentalBERT~\cite{zhai2024chinese} model excelled, achieving an F1 score of 88.39\%.
In this task, the advantage of Chinese MentalBERT over MacBERT\cite{cui2021pre} is evident, achieving an F1-score that is 4.8\% points higher.

From these two tasks, it is apparent that the experimental models perform well in the high-low suicide classification task, but struggle in the fine-grained classification task, likely due to category imbalance. Consequently, we continued to explore various data augmentation strategies to address this issue.

\begin{table}[h]
\centering
\caption{Performance of seven pre-trained models on two suicide classification tasks. Precision, recall, and F1-score were reported in weighted average.}
\resizebox{1\linewidth}{!}{
\begin{tabular}{llccc} 
\hline
\textbf{Tasks}                       & \textbf{Models}              & \textbf{Precision} & \textbf{Recall}  & \textbf{F1}        \\ 
\hline
\multirow{7}{*}{Fine-grained      } & BERT                        & 48.95\%            & 50.80\%          & 46.16\%            \\
& RoBERTa                     & 38.03\%            & 42.80\%          & 37.11\%            \\
& ELECTRA                     & 30.35\%            & 42.40\%          & 34.91\%            \\
& MacBERT                     & 50.17\%            & 53.60\%          & 50.59\%            \\
& NeZha                       & 47.61\%            & 50.40\%          & 45.72\%            \\
& ERNIE 3.0                   & 47.90\%            & 51.20\%          & 47.36\%            \\
                                    & \textbf{Chinese MentalBERT} & \textbf{52.81\%}   & \textbf{54.80\%} & \textbf{50.89\%}   \\ 
\hline
\multirow{7}{*}{High-low      }     & BERT                        & 87.68\%            & 87.60\%          & 87.61\%            \\
& RoBERTa                     & 80.02\%            & 80.00\%          & 79.97\%            \\
                                    & ELECTRA                     & 81.21\%            & 81.20\%          & 81.20\%            \\
                                    & MacBERT                     & 83.60\%            & 83.60\%          & 83.59\%            \\
                                    & NeZha                       & 84.86\%            & 84.80\%          & 84.81\%            \\
                                    & ERNIE 3.0                   & 86.42\%            & 86.40\%          & 86.39\%            \\
                                    & \textbf{Chinese MentalBERT} & \textbf{88.41\%}   & \textbf{88.40\%} & \textbf{88.39\% }  \\
\hline
\end{tabular}
}
\label{tab:result}
\end{table}

\subsection{Model performance after data augmentation}
As shown in Table~\ref{tab:dataset} and Table~\ref{tab:result}, the fine-grained suicide risk classification task has the challenge of class imbalance, and so we continued to experiment with several data augmentation methods in an attempt to address this issue. 
The model performance after data augmentation are detailed in Table~\ref{tab:result_aug}. 

\begin{table*}
\centering
\caption{Weighted F1 scores for each model after data augmentation and comparison to baseline in the fine-grained suicide risk classification task. Three augmentation strategies are implemented: round-trip translation (RT), synonym substitution (SR), and LLM generation (LLM-G). The most effective training strategy for each model is highlighted in bold.}
\resizebox{\linewidth}{!}
{
\begin{tabular}{cccccc} 
\toprule
\textbf{Model}              & \multicolumn{1}{c}{\textbf{Baseline}} & \textbf{RT}            & \textbf{SR}            & \multicolumn{1}{c}{\textbf{LLM}} & \multicolumn{1}{c}{\textbf{2$\times$LLM}}  \\ 
\bottomrule
BERT                        & 46.16\%                               & 47.05\%(+0.89\%)          & 47.70\%(+1.54\%)          & \textbf{48.94\%(+2.78\%)}              & 48.49\%(+2.33\%)                        \\
RoBERTa                     & 37.11\%                               & 38.92\%(+1.81\%)          & 38.10\%(+0.99\%)          & 38.43\%(+1.32\%)                       & \textbf{41.91\%(+4.80\%)}               \\
ELECTRA                     & 34.91\%                               & 32.24\%(-2.67\%)          & 35.52\%(+0.61\%)          & 38.15\%(+3.24\%)                       & \textbf{39.69\%(+4.78\%)}               \\
MacBERT                     & \textbf{50.59\%}                      & 47.20\%(-3.39\%)          & 45.01\%(-5.58\%)          & 46.44\%(-4.15\%)                       & 48.38\%(-2.21\%)                        \\
NeZha                       & 45.72\%                               & 46.80\%(+1.08\%)          & \textbf{49.66\%(+3.94\%)} & 48.84\%(+3.12\%)                       & 48.81\%(+3.09\%)                        \\
ERNIE 3.0                   & 47.36\%                               & 48.26\%(+0.90\%)          & 49.44\%(+2.08\%)          & 49.22\%(+1.86\%)                       & \textbf{51.22\%(+3.86\%) }              \\
\textbf{Chinese MentalBERT} & 50.89\%                               & \textbf{55.54\%(+4.65\%)} & 53.46\%(+2.57\%)          & 51.05\%(+0.16\%)                       & 53.97\%(+3.08\%)                        \\
\toprule
\end{tabular}
}
\label{tab:result_aug}
\end{table*}

From the experimental results we can found that each strategy has its merits and limitations depending on the model architecture. RT generally shows positive results, particularly excelling with the Chinese MentalBERT model where it led to a significant improvement of +4.65\%. However, it did not universally enhance performance, as seen with ELECTRA, where it resulted in a decrement. SR was notably effective for NeZha, achieving the highest performance boost of +3.94\% for this model, but it caused a substantial performance decline in MacBERT (-5.58\%). This indicates that SR's effectiveness is highly contingent on how each model processes semantic variations. LLM-G provided consistent, moderate improvements across most models, with the notable exception of MacBERT, reinforcing its reliability but also highlighting that the magnitude of improvement is generally less dramatic compared to other strategies. 
Due to the consistently improving performance of LLM-G, we continued to employ the double maximal category augmentation strategy (2$\times$LLM) to investigate the effects of doubling the augmented data on the model.

When the LLM augmentation volume was doubled, there was a marked improvement in model performance for models like RoBERTa, ELECTRA, and ERNIE 3.0, suggesting that a higher volume of LLM-generated data can be beneficial, particularly for models that already showed positive responses to the initial LLM-G strategy. For RoBERTa and ELECTRA, the increase in data volume led to the highest improvements observed (+4.80\% and +4.78\%, respectively), indicating that these models may benefit from larger training datasets. Conversely, NeZha and BERT did not see the best results with this increased volume, suggesting there may be an optimal point of augmentation that varies by model. For MacBERT and Chinese MentalBERT, doubling the LLM volume did not yield improvements that surpassed other strategies, with Chinese MentalBERT performing best under RT and MacBERT still not reaching its baseline performance.

The experimental result highlights that while LLM-G offers a reliable base for performance improvement, increasing the volume of data augmentation with strategies like 2$\times$LLM can provide benefits for certain models, especially where initial improvements were promising. However, the varying effectiveness of RT, SR, and LLM strategies across different models suggests that data augmentation approaches must be tailored to individual model architectures and specific tasks to optimize outcomes.

\section{Discussion} 
\label{sec:Discussion}

In this study, we employed the GPT-4 to assist in annotating suicide risk levels. The model encountered several challenges in accurately understanding the data. First, it struggled to differentiate various levels of severity, such as distinguishing between the intensity of suicidal ideation and the degree of existential pain. Additionally, it often misinterpreted narratives that mentioned potential suicidal actions in the future as indications of a definite suicide plan. The brevity of the data further complicated the model's ability to understand the true semantic meaning of sentences, especially when it came to internet slang. For instance, phrases like 
\begin{CJK}{UTF8}{gbsn}
``约死滴滴 (English translation: Beep, beep, calling for a suicide pact.)'', 
\end{CJK} a type of internet slang, were sometimes incorrectly assumed to refer to a specific method of suicide.

In the field of fine-grained suicide risk classification, several challenges persist, particularly the need to mitigate the subjectivity of annotators. Additionally, the collection and labeling of large datasets are required to address issues of data imbalance. While data augmentation techniques have been effective in enhancing model performance, there remains room for improvement. One possible solution is to increase semantic diversity or incorporate internet slang to better reflect the actual language used online. Furthermore, models need to be trained to detect subtle emotional and semantic cues in text, thereby enhancing their capability for fine-grained emotional understanding. In psychology, binary classification tasks are useful for quickly identifying high-risk individuals. Deep learning models have shown strong capabilities in robust classification, making them a practical solution in contemporary settings.

\section{Conclusion}
\label{sec:conclusion}
This study introduces the {SOS-1K} dataset, which categorizes suicide risk by urgency levels. We evaluated seven pre-trained language models on this dataset for both fine-grained and high-low suicide classification tasks.
Given the class imbalance in fine-grained classification tasks, we explored three data augmentation methods: synonym replacement, round-trip translation, and data generation by LLM. 
The results indicate that data augmentation enhances model performance, with data generated by LLM generally contributing positively. Notably, the Chinese MentalBERT model demonstrated superior performance, achieving a 55.54\% weighted F1-score in the fine-grained classification task and an 88.39\% F1-score in the binary high-low suicide risk classification task. These findings underscore the advantages of domain-specific pre-trained models.
The datasets, source code, and augmented data are publicly accessible. This research establishes a foundation for future studies and provides strategies for social media based suicide risk interventions.

\bibliographystyle{IEEEtran} 
\bibliography{refs}

\begin{thebibliography}{10}
\providecommand{\url}[1]{#1}
\csname url@samestyle\endcsname
\providecommand{\newblock}{\relax}
\providecommand{\bibinfo}[2]{#2}
\providecommand{\BIBentrySTDinterwordspacing}{\spaceskip=0pt\relax}
\providecommand{\BIBentryALTinterwordstretchfactor}{4}
\providecommand{\BIBentryALTinterwordspacing}{\spaceskip=\fontdimen2\font plus
\BIBentryALTinterwordstretchfactor\fontdimen3\font minus \fontdimen4\font\relax}
\providecommand{\BIBforeignlanguage}[2]{{%
\expandafter\ifx\csname l@#1\endcsname\relax
\typeout{** WARNING: IEEEtran.bst: No hyphenation pattern has been}%
\typeout{** loaded for the language `#1'. Using the pattern for}%
\typeout{** the default language instead.}%
\else
\language=\csname l@#1\endcsname
\fi
#2}}
\providecommand{\BIBdecl}{\relax}
\BIBdecl

\bibitem{world2022world}
W.~H. Organization \emph{et~al.}, ``World mental health report: transforming mental health for all,'' 2022.

\bibitem{keles2020systematic}
B.~Keles, N.~McCrae, and A.~Grealish, ``A systematic review: the influence of social media on depression, anxiety and psychological distress in adolescents,'' \emph{International journal of adolescence and youth}, vol.~25, no.~1, pp. 79--93, 2020.

\bibitem{chen2020treehole}
P.~Chen, Y.~Qian, Z.~Huang, C.~Zhao, Z.~Liu, and B.~Yang, ``Negative emotional characteristics of weibo ``tree hole''' users,'' \emph{Zhongguo Xinliweisheng Zazhi}, vol.~5, pp. 437--444, 2020.

\bibitem{yang2021suicide}
B.~X. Yang, L.~Xia, L.~Liu, W.~Nie, Q.~Liu, X.~Y. Li, M.~Q. Ao, X.~Q. Wang, Y.~D. Xie, Z.~Liu \emph{et~al.}, ``A suicide monitoring and crisis intervention strategy based on knowledge graph technology for ``tree hole'' microblog users in {China},'' \emph{Frontiers in psychology}, vol.~12, p. 674481, 2021.

\bibitem{yimeng2020aigives}
Z.~Yimeng and L.~Kun, ``{AI} gives potential suicides pause for thought,'' \emph{China Daily}, 2020.

\bibitem{coppersmith2018natural}
G.~Coppersmith, R.~Leary, P.~Crutchley, and A.~Fine, ``Natural language processing of social media as screening for suicide risk,'' \emph{Biomedical informatics insights}, vol.~10, p. 1178222618792860, 2018.

\bibitem{fu2021distant}
G.~Fu, C.~Song, J.~Li, Y.~Ma, P.~Chen, R.~Wang, B.~X. Yang, and Z.~Huang, ``Distant supervision for mental health management in social media: suicide risk classification system development study,'' \emph{Journal of medical internet research}, vol.~23, no.~8, p. e26119, 2021.

\bibitem{renjith2022ensemble}
S.~Renjith, A.~Abraham, S.~B. Jyothi, L.~Chandran, and J.~Thomson, ``An ensemble deep learning technique for detecting suicidal ideation from posts in social media platforms,'' \emph{Journal of King Saud University-Computer and Information Sciences}, vol.~34, no.~10, pp. 9564--9575, 2022.

\bibitem{gorai2024bert}
J.~Gorai and D.~K. Shaw, ``A {BERT}-encoded ensembled {CNN} model for suicide risk identification in social media posts,'' \emph{Neural Computing and Applications}, pp. 1--16, 2024.

\bibitem{wang2021medical}
R.~Wang, B.~X. Yang, Y.~Ma, P.~Wang, Q.~Yu, X.~Zong, Z.~Huang, S.~Ma, L.~Hu, K.~Hwang \emph{et~al.}, ``Medical-level suicide risk analysis: A novel standard and evaluation model,'' \emph{IEEE Internet of Things Journal}, vol.~8, no.~23, pp. 16\,825--16\,834, 2021.

\bibitem{shorten2021text}
C.~Shorten, T.~M. Khoshgoftaar, and B.~Furht, ``Text data augmentation for deep learning,'' \emph{Journal of big Data}, vol.~8, no.~1, p. 101, 2021.

\bibitem{10.1162/tacl_a_00542}
\BIBentryALTinterwordspacing
J.~Chen, D.~Tam, C.~Raffel, M.~Bansal, and D.~Yang, ``{An Empirical Survey of Data Augmentation for Limited Data Learning in NLP},'' \emph{Transactions of the Association for Computational Linguistics}, vol.~11, pp. 191--211, 03 2023. [Online]. Available: \url{https://doi.org/10.1162/tacl\_a\_00542}
\BIBentrySTDinterwordspacing

\bibitem{zhao2023survey}
W.~X. Zhao, K.~Zhou, J.~Li, T.~Tang, X.~Wang, Y.~Hou, Y.~Min, B.~Zhang, J.~Zhang, Z.~Dong \emph{et~al.}, ``A survey of large language models,'' \emph{arXiv preprint arXiv:2303.18223}, 2023.

\bibitem{kaddour2023challenges}
J.~Kaddour, J.~Harris, M.~Mozes, H.~Bradley, R.~Raileanu, and R.~McHardy, ``Challenges and applications of large language models,'' \emph{arXiv preprint arXiv:2307.10169}, 2023.

\bibitem{he2023towards}
T.~He, G.~Fu, Y.~Yu, F.~Wang, J.~Li, Q.~Zhao, C.~Song, H.~Qi, D.~Luo, H.~Zou \emph{et~al.}, ``Towards a psychological generalist {AI}: A survey of current applications of large language models and future prospects,'' \emph{arXiv preprint arXiv:2312.04578}, 2023.

\bibitem{liu2023chatcounselor}
J.~M. Liu, D.~Li, H.~Cao, T.~Ren, Z.~Liao, and J.~Wu, ``Chatcounselor: A large language models for mental health support,'' \emph{arXiv preprint arXiv:2309.15461}, 2023.

\bibitem{fu2023enhancing}
G.~Fu, Q.~Zhao, J.~Li, D.~Luo, C.~Song, W.~Zhai, S.~Liu, F.~Wang, Y.~Wang, L.~Cheng \emph{et~al.}, ``Enhancing psychological counseling with large language model: A multifaceted decision-support system for non-professionals,'' \emph{arXiv preprint arXiv:2308.15192}, 2023.

\bibitem{zhou2023identifying}
W.~Zhou, L.~C. Prater, E.~V. Goldstein, S.~J. Mooney \emph{et~al.}, ``Identifying rare circumstances preceding female firearm suicides: validating a large language model approach,'' \emph{JMIR mental health}, vol.~10, no.~1, p. e49359, 2023.

\bibitem{ayers2023comparing}
J.~W. Ayers, A.~Poliak, M.~Dredze, E.~C. Leas, Z.~Zhu, J.~B. Kelley, D.~J. Faix, A.~M. Goodman, C.~A. Longhurst, M.~Hogarth \emph{et~al.}, ``Comparing physician and artificial intelligence chatbot responses to patient questions posted to a public social media forum,'' \emph{JAMA internal medicine}, vol. 183, no.~6, pp. 589--596, 2023.

\bibitem{qi2023supervised}
H.~Qi, Q.~Zhao, J.~Li, C.~Song, W.~Zhai, L.~Dan, S.~Liu, Y.~J. Yu, F.~Wang, H.~Zou \emph{et~al.}, ``Supervised learning and large language model benchmarks on mental health datasets: Cognitive distortions and suicidal risks in chinese social media,'' 2023.

\bibitem{zhai2024chinese}
W.~Zhai, H.~Qi, Q.~Zhao, J.~Li, Z.~Wang, H.~Wang, B.~X. Yang, and G.~Fu, ``Chinese {MentalBERT}: Domain-adaptive pre-training on social media for chinese mental health text analysis,'' \emph{arXiv preprint arXiv:2402.09151}, 2024.

\bibitem{xiao2024decoding}
Y.~Xiao, K.~Bi, P.~S.-F. Yip, J.~Cerel, T.~T. Brown, Y.~Peng, J.~Pathak, and J.~J. Mann, ``Decoding suicide decedent profiles and signs of suicidal intent using latent class analysis,'' \emph{JAMA psychiatry}, 2024.

\bibitem{cowen2012shorter}
P.~Cowen, P.~Harrison, and T.~Burns, \emph{Shorter Oxford textbook of psychiatry}.\hskip 1em plus 0.5em minus 0.4em\relax Oxford University Press, USA, 2012.

\bibitem{Huang2019WebBasedIA}
Z.~Huang, Q.~Hu, J.~Gu, J.~Yang, Y.~Feng, and G.~Wang, ``Web-based intelligent agents for suicide monitoring and early warning,'' \emph{China Digital Medicine}, vol.~14, no.~3, pp. 2--6, 2019.

\bibitem{kenton2019bert}
J.~D. M.-W.~C. Kenton and L.~K. Toutanova, ``{BERT}: Pre-training of deep bidirectional transformers for language understanding,'' in \emph{Proceedings of naacL-HLT}, vol.~1, 2019, p.~2.

\bibitem{vaswani2017attention}
A.~Vaswani, N.~Shazeer, N.~Parmar, J.~Uszkoreit, L.~Jones, A.~N. Gomez, {\L}.~Kaiser, and I.~Polosukhin, ``Attention is all you need,'' \emph{Advances in neural information processing systems}, vol.~30, 2017.

\bibitem{liu2019roberta}
Y.~Liu, M.~Ott, N.~Goyal, J.~Du, M.~Joshi, D.~Chen, O.~Levy, M.~Lewis, L.~Zettlemoyer, and V.~Stoyanov, ``{RoBERTa}: A robustly optimized bert pretraining approach,'' \emph{arXiv preprint arXiv:1907.11692}, 2019.

\bibitem{clark2020electra}
K.~Clark, M.-T. Luong, Q.~V. Le, and C.~D. Manning, ``{ELECTRA}: Pre-training text encoders as discriminators rather than generators,'' \emph{arXiv preprint arXiv:2003.10555}, 2020.

\bibitem{cui2021pre}
Y.~Cui, W.~Che, T.~Liu, B.~Qin, and Z.~Yang, ``Pre-training with whole word masking for {Chinese} {BERT},'' \emph{IEEE/ACM Transactions on Audio, Speech, and Language Processing}, vol.~29, pp. 3504--3514, 2021.

\bibitem{wei2019nezha}
J.~Wei, X.~Ren, X.~Li, W.~Huang, Y.~Liao, Y.~Wang, J.~Lin, X.~Jiang, X.~Chen, and Q.~Liu, ``Nezha: Neural contextualized representation for chinese language understanding,'' \emph{arXiv preprint arXiv:1909.00204}, 2019.

\bibitem{sun2021ernie}
Y.~Sun, S.~Wang, S.~Feng, S.~Ding, C.~Pang, J.~Shang, J.~Liu, X.~Chen, Y.~Zhao, Y.~Lu \emph{et~al.}, ``{ERNIE} 3.0: Large-scale knowledge enhanced pre-training for language understanding and generation,'' \emph{arXiv preprint arXiv:2107.02137}, 2021.

\bibitem{chen2023empirical}
J.~Chen, D.~Tam, C.~Raffel, M.~Bansal, and D.~Yang, ``An empirical survey of data augmentation for limited data learning in {NLP},'' \emph{Transactions of the Association for Computational Linguistics}, vol.~11, pp. 191--211, 2023.

\bibitem{kolomiyets2011model}
O.~Kolomiyets, S.~Bethard, and M.-F. Moens, ``Model-portability experiments for textual temporal analysis,'' in \emph{Proceedings of the 49th annual meeting of the association for computational linguistics: human language technologies}, vol.~2.\hskip 1em plus 0.5em minus 0.4em\relax ACL; East Stroudsburg, PA, 2011, pp. 271--276.

\bibitem{aiken2010efficacy}
M.~Aiken and M.~Park, ``The efficacy of round-trip translation for {MT} evaluation,'' \emph{Translation Journal}, vol.~14, no.~1, pp. 1--10, 2010.

\bibitem{salton1975vector}
G.~Salton, A.~Wong, and C.-S. Yang, ``A vector space model for automatic indexing,'' \emph{Communications of the ACM}, vol.~18, no.~11, pp. 613--620, 1975.

\bibitem{Synonyms:hain2017}
\BIBentryALTinterwordspacing
H.~Y.~X. Hai Liang~Wang. (2017) Chinese synonym toolkit: Synonyms. [Online]. Available: \url{https://github.com/chatopera/Synonyms}
\BIBentrySTDinterwordspacing

\bibitem{BaiduTranslateAPI}
{Baidu, Inc.}, ``Baidu translate api,'' \url{https://api.fanyi.baidu.com/}, accessed: 2024-02-21.

\bibitem{achiam2023gpt}
J.~Achiam, S.~Adler, S.~Agarwal, L.~Ahmad, I.~Akkaya, F.~L. Aleman, D.~Almeida, J.~Altenschmidt, S.~Altman, S.~Anadkat \emph{et~al.}, ``{GPT}-4 technical report,'' \emph{arXiv preprint arXiv:2303.08774}, 2023.

\bibitem{zoufan}
\BIBentryALTinterwordspacing
ZouFan, ``Sina weibo "{Zoufan}" comment,'' 2023. [Online]. Available: \url{https://www.weibo.com/xiaofan116?is_all=1}
\BIBentrySTDinterwordspacing

\bibitem{paszke2019pytorch}
A.~Paszke, S.~Gross, F.~Massa, A.~Lerer, J.~Bradbury, G.~Chanan, T.~Killeen, Z.~Lin, N.~Gimelshein, L.~Antiga \emph{et~al.}, ``{PyTorch}: An imperative style, high-performance deep learning library,'' \emph{Advances in neural information processing systems}, vol.~32, 2019.

\bibitem{kingma2015adam}
D.~P. Kingma and J.~Ba, ``Adam: A method for stochastic optimization,'' in \emph{3rd International Conference on Learning Representations, ICLR 2015 - Conference Track Proceedings}, San Diego, CA, USA, 2015.

\end{thebibliography}

\end{document}